\documentclass[journal]{IEEEtran}

\usepackage{times}
\usepackage{epsfig}
\usepackage{graphicx}
\usepackage{amsmath}
\usepackage{amssymb}
\usepackage[linesnumbered,ruled,vlined]{algorithm2e}
\usepackage{comment,color,soul}
\usepackage{color, colortbl}
\usepackage{makecell}
\setcellgapes{1.5pt}

\definecolor{LightCyan}{rgb}{0.88,1,1}
\definecolor{Gray}{gray}{0.9}

\ifCLASSINFOpdf
\else
\fi
\begin{document}
%
\title{Towards Low-Latency Energy-Efficient Deep SNNs via Attention-Guided Compression}
%
%
%

\author{Souvik~Kundu,~\IEEEmembership{Student Member,~IEEE,}
        Gourav~Datta,~\IEEEmembership{Student Member,~IEEE,}
        Massoud Pedram,~\IEEEmembership{Fellow,~IEEE,}
        Peter~A.~Beerel,~\IEEEmembership{Senior Member,~IEEE}
\thanks{S. Kundu, G. Datta, M. Pedram and P. A. Beerel are with the Department of Electrical and Computer Engineering, University of Southern California, Los Angeles,
CA, 90089 USA e-mail: \{souvikku, gdatta, pedram, pabeerel\}@usc.edu.}
}

%
%

\markboth{Towards Low-Latency Energy-efficient Deep SNNs via AGC}%
{Shell \MakeLowercase{\textit{et al.}}: Bare Demo of IEEEtran.cls for IEEE Journals}
%



\maketitle

\begin{abstract}
Deep spiking neural networks (SNNs) have emerged as a potential alternative to traditional deep learning frameworks, due to their promise to provide increased compute efficiency on event-driven neuromorphic hardware. However, to perform well on complex vision applications, most SNN training frameworks yield large inference 
latency which translates to increased spike activity and reduced energy efficiency. 
Hence, minimizing average spike activity while preserving accuracy in deep SNNs remains a significant challenge and opportunity.
This paper presents a non-iterative SNN training technique that achieves ultra-high compression with reduced spiking activity 
while maintaining high inference accuracy.
In particular, our framework first uses the attention-maps of an uncompressed meta-model to yield compressed ANNs. This step can be tuned to support both irregular and structured channel pruning to leverage computational benefits over a broad range of platforms. The framework then performs sparse-learning-based supervised SNN training using direct inputs. During the training, it jointly optimizes the SNN weight, threshold, and leak parameters to drastically minimize the 
number of time steps required while retaining compression.
To evaluate the merits of our approach, we performed experiments with variants of VGG and ResNet, on both CIFAR-10 and CIFAR-100, and VGG16 on Tiny-ImageNet. The SNN models generated through the proposed technique yield state-of-the-art compression ratios of up to $33.4 \times$ with no significant drop in accuracy compared to baseline unpruned counterparts. Compared to existing SNN pruning methods, we achieve up to $8.3 \times$ higher compression with improved accuracy. In particular, we show that our compressed VGG16 models on CIFAR-10 have up to $\mathord{\sim}98 \times$ and $\mathord{\sim}47 \times$ better compute energy efficiency compared to uncompressed and iso-parameter ANNs, respectively.   
\end{abstract}


%
\IEEEpeerreviewmaketitle

\section{Introduction}
%
%
%
%
\label{sec:intro}
\IEEEPARstart{A}{rtificial} neural networks (ANNs) have shown enormous success in various applications ranging from object recognition \cite{ szegedy2015going, he2016deep} and detection \cite{redmon2017yolo9000} to image segmentation \cite{tao2018image}. However, their large model size and costly energy budget have limited their deployment to resource constrained IoT applications.  Moreover, user data-privacy have also fueled the need for on-chip intelligence in resource-constrained edge devices.
To that end, spiking neural networks (SNNs), originally inspired by the functionality of biological neurons \cite{mainen1995reliability}, have gained popularity due to their promise for low-power machine learning \cite{neuro_frontiers,spike_ratecoding}. 
The underlying SNN hardware leverages a binary spike-based sparse processing over a fixed number of time steps via accumulate (AC) operations, consuming much lower power than multiply-accumulate (MAC) operations that dominate computations in ANNs \cite{farabet2012comparison}. However, the non-differentiability of binary spikes has made effective gradient-descent-based training elusive, and consequently, SNNs have generally performed relatively poorly in complex vision tasks.  Various recent research efforts have started to address this issue through different forms of spike-driven supervised \cite{connor_sensf, lee_dsnn, lee_stdp, bellec_2018long, lee2019enabling}, conversion-based indirect supervised \cite{dsnn_conversion_ijcnn, dsnn_conversion_abhronilfin, diehl2016conversion}, and unsupervised learning \cite{gopal_restoch,rathi2017stdp}. More recently, a hybrid training technique, which first trains a specifically-designed ANN, then converts it to an SNN, and finally fine-tunes it for a few epochs \cite{rathi2020iclr}, has yielded state of the art results.
Compared to the ANN-to-SNN conversion based approach \cite{dsnn_conversion_abhronilfin}, this hybrid ANN-SNN training reduces the number of required inference time steps by $\mathord{\sim}10 \times$, decreasing spiking activity and thus energy consumption. 

\begin{figure}[!t]
\includegraphics[width=0.45\textwidth]{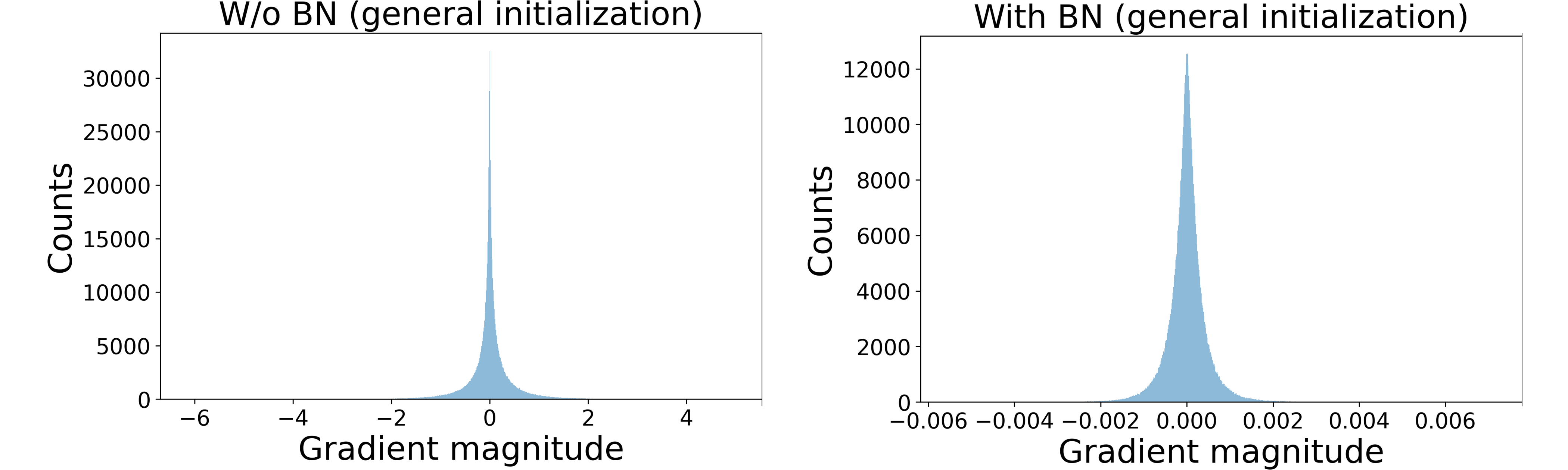}
\centering
   \caption{Histogram of the gradients for CONV layer 7 of VGG16 with target density of 0.4 at an early stage of training (after 10 epochs) to classify CIFAR-10.}
\label{fig:grad_explode}
\vspace{-6mm}
\end{figure}

Model compression including pruning \cite{han2015learning, kundu2021dnr, dettmers2019sparse} and quantization \cite{ courbariaux2016binarized} have mitigated the energy-consumption issue in ANNs. In particular, {\em irregular pruning} forces individual insignificant weights to be zero, yielding sparse weight tensors that can lower energy consumption but require dedicated hardware to yield inference speedup. In contrast, {\em structured pruning} retains the original convolutional structure by pruning at the level of channels or layers and thus can help realize lower power and inference speedup without dedicated hardware support \cite{liu2018rethinking}. Unfortunately, their application to SNNs has remained an arduous task. 
In particular, for hybrid training or training based on ANN-to-SNN conversion, it has been recommended that the ANN models not have batch-normalization (BN) layers \cite{dsnn_conversion_abhronilfin, rathi2020iclr}.  This architectural constraint is particularly important as BN layers play a key role in training loss convergence \cite{bjorck2018understanding} and their absence makes achieving significant compression without a large performance drop challenging. Moreover, it has been observed that the spike coding of SNNs makes their accuracy very sensitive to model compression \cite{deng2019comprehensive}.

Among the handful of works on SNN compression using pruning \cite{soft_pruning, rathi2017stdp}, most are limited to shallow networks on small datasets such as MNIST. A recent effort \cite{deng2019comprehensive} has combined spatio-temporal backpropagation (STBP) with alternating direction method of multipliers (ADMM) to prune SNNs via spike-based training. However, due to backpropagation through time, SNN training procedures are extremely memory intensive and have long training times \cite{rathi2020iclr}. 
Moreover, achieving optimum performance with ADMM demands either hand-tuning or a separate evaluation technique to find the per-layer target parameter density, which itself is a tedious iterative procedure often requiring expert insight into the model \cite{liu2020autocompress}. Due to this memory intensive nature any iterative pruning \cite{frankle2018lottery} becomes prohibitively costly in SNN domain. 

 Recently developed sparse-learning (SL) method \cite{dettmers2019sparse, kundu2021dnr} has emerged as  a promising compression solution for ANNs because it can achieve high compression in a single training iteration with better accuracy than many other approaches \cite{han2015learning, zhang2018systematic, he2018amc}. Moreover, unlike ADMM this method does not require per-layer target parameter density. However, this non-iterative strategy suffers from non-convergence in BN-free deep ANN compression, which is in turn caused by the gradient-explosion issue as illustrated in Fig. \ref{fig:grad_explode}.

This paper presents a non-iterative compression technique for deep SNNs referred as $\textit{attention-guided}$ compression (AGC). In particular, our sparse-learning strategy uses attention-maps of an unpruned pre-trained meta model (Fig. \ref{fig:agc_framework}) to mitigate non-convergence of the BN-free ANN and guide the compression. This approach is different from the idea of distillation \cite{zagoruyko2016paying, hinton2015distilling}, because the meta-model in our approach can be of lower complexity than the model to be compressed and does not use the KL-divergence loss between the models. 
In our approach we first compress an ANN model specifically-designed for SNN conversion and then apply the ANN-to-SNN conversion technique \cite{dsnn_conversion_abhronilfin}. After conversion to an SNN, to reduce the number of time steps required, we extend the hybrid SNN training strategy by using a compression knowledge-driven SL-based SNN training approach. AGC only requires a global parameter density and thus can avoid iterative tuning of per-layer density which is necessary for ADMM. Our initial results with AGC, presented in \cite{Kundu2021WACV}, show negligible performance degradation compared to the uncompressed baseline models.

\begin{figure}[!t]
\includegraphics[width=0.49\textwidth]{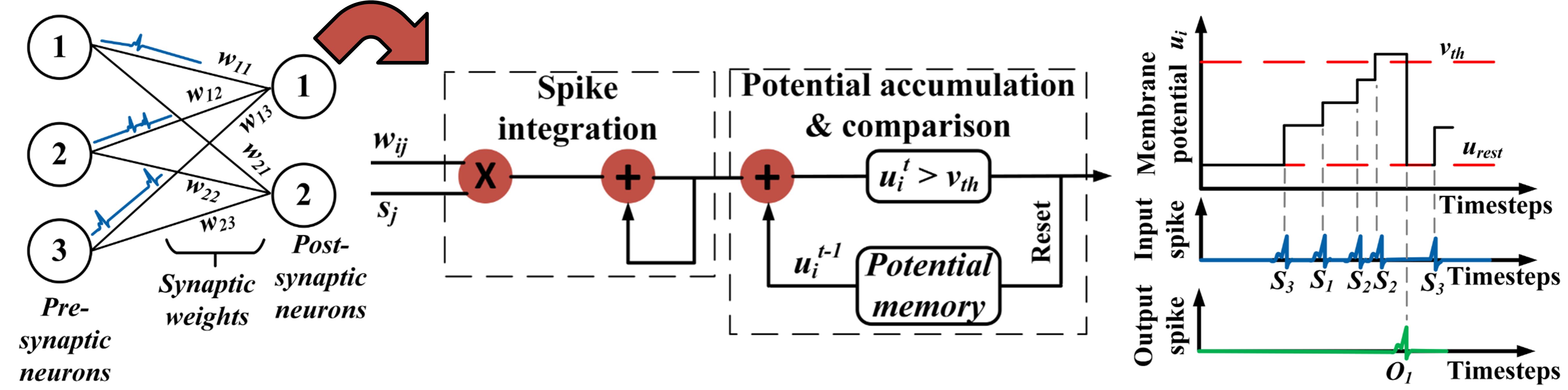}
\centering
   \caption{SNN fundamental operations.}
\label{fig:SNN_fundamentals}
\vspace{-4mm}
\end{figure}

\noindent


We further extend AGC framework to support structured pruning, namely {\em channel pruning}, enabling benefits on a broader class of compute platforms. Structured AGC can yield models with reduced effective convolutional channel width and thus has the potential to significantly speed up inference. To the best of our knowledge, we are the first to propose a non-iterative SNN compression that supports both irregular and channel pruning. 
Moreover, to further reduce the SNN training time, thereby reducing the inference latency and compute energy consumption, we extend our hybrid SL framework to support direct-coded input and allow both the firing threshold and leak parameters to be trainable \cite{rathi2020diet} (along with the non-zero weights). Our experimental results show that compared to our initial AGC framework \cite{Kundu2021WACV} which only supported rate-coded inputs, this enhanced AGC can lead up to $15\times$ reduced inference latency and $\mathord{\sim}4.5 \times$ lower inference compute energy. 

We analyze the benefits of our enhanced AGC using VGG \cite{simonyan2014very} and ResNet \cite{he2016deep} model variants on CIFAR-10  and CIFAR-100 \cite{krizhevsky2009learning}, and using VGG16 on Tiny-ImageNet \cite{hansen2015tiny}. We  benchmark the model performances by using both the standard metric of average spike count per layer and a novel metric that captures the compute efficiency for both the direct and rate-coded input driven sparse model \cite{Kundu2021WACV}.
Furthermore, after analyzing the compressed model performance and layer wise spiking activity, we provide observations to give insight and help guide deep SNN training to yield better performance both in terms of FLOPs and accuracy.

The remainder of this paper is structured as follows. In Section \ref{sec:back} we present necessary background and related work. Section \ref{sec:snn_agc} and \ref{sec:expt} describe our SNN compression technique and provide detailed experimental evaluations, respectively. Finally the paper concludes in Section \ref{sec:concl}.

\section{Background and Related Work}
\label{sec:back}
\subsection{SNN Fundamentals}

In ANNs, inference is performed in a single feed-forward pass through the network. 
In SNNs, in contrast, network of neurons communicate through a sequence
of binary spikes over a defined number of time steps $T$, referred to as the SNN's
inference latency.
The spiking dynamics in an SNN layer is characterized with either the Integrate-Fire (IF) \cite{lu2020exploring} or Leaky-Integrate-Fire (LIF) \cite{leefin2020} neuron model. The iterative version of the LIF neuron dynamics is defined by the following differential equation
\vspace{-2mm}
\begin{align}
    u_i^{t+1} = (1-\frac{dt}{\tau})u_i^t + \frac{dt}{\tau}I 
\label{eq:neuron_diff}
\end{align}
\vspace{-2mm}

\noindent
where $u_i^{t+1}$ represents the membrane potential of the $i^{th}$ neuron at time step $t+1$, $\tau$ is a time constant, and $I$ is the input from one of possibly many pre-synaptic neurons. To evaluate this model in discrete time \cite{wu2018spatio}, the model can be modified as
\vspace{-1mm}
\begin{align}
    u_i^{t+1} &= \lambda u_i^t + \sum_{j}w_{ij}O_j^{t} - v_{th}O_i^t\\
    O_i^{t}   &=
    \begin{cases}
    1, & \text{if } u_i^t>v_{th}\\
    0, & \text{otherwise}
    \end{cases}
\label{eq:neuron_discrete}
\end{align}
\vspace{-2.7mm}

\noindent
where the decay factor $(1-{dt}/{\tau})$ 
is replaced by the term $\lambda$, which is set to 1 for IF and less than 1 for LIF. Here, $O_i^{t}$ and $O_j^{t}$ represent the output spikes of current neuron $i$ and one of its pre-synaptic neurons $j$, respectively. $w_{ij}$ represents the weight between the two and $v_{th}$ is the firing threshold of the current layer. The result of the inference is obtained by comparing the total number of spikes generated by each output neuron over $T$ time steps. Training SNNs is challenging because 1) exact gradients for the binary spike trains are undefined, forcing the use of approximate gradients, and 2) the training complexity scales with the number of time steps $T$, which can be large.

\begin{figure*}[!t]
\includegraphics[width=0.92\textwidth]{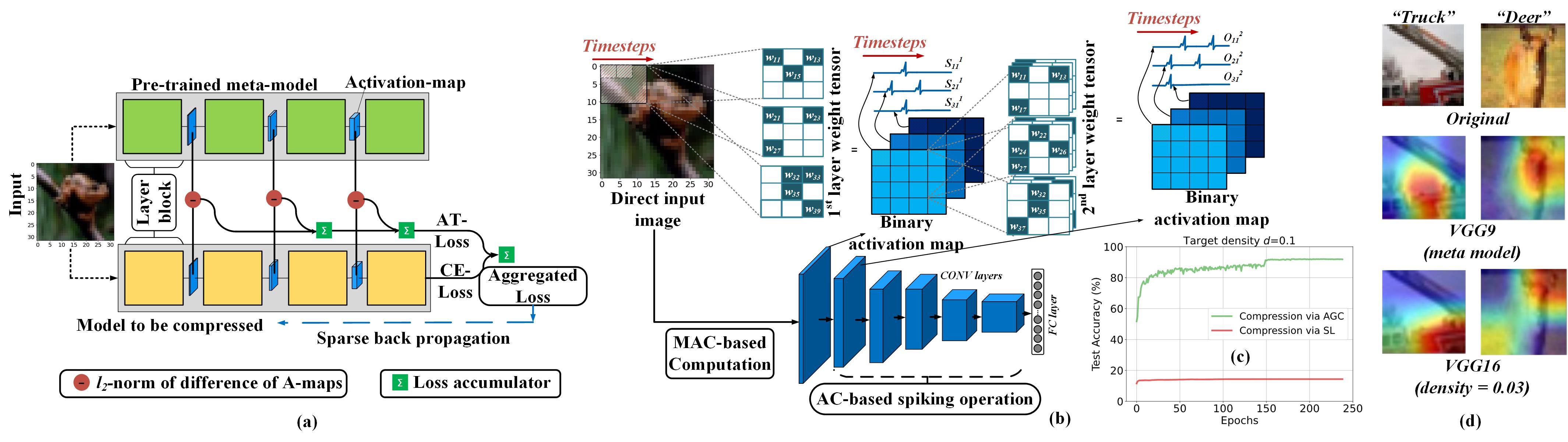}
\centering
   \caption{Two major training stages of the proposed scheme: (a) ANN training using attention-guided compression (AGC) and (b) Sparse-learning based SNN training using surrogate gradient-based training. Because we use direct input coding for the input, the first layer of the SNN uses MAC-based computation. The second and subsequent convolution layers only require accumulate operations. (c) Plot of test accuracy versus epochs for ResNet12 on CIFAR-10 for model compressed using AGC with VGG9 chosen as $\Psi_m$. (d) Activation map of the last convolution layer of the $\Psi_m$ (VGG9) and corresponding $\Psi_c$ (VGG16) with target density, $d$ = 0.03 for two images selected from CIFAR-10.}
\label{fig:agc_framework}
\vspace{-5mm}
\end{figure*}

\subsection{SNN Training Background}
\subsubsection{ANN-to-SNN Conversion}
\label{subsec:ann_snn_conv}
The ANN-to-SNN conversion-based training algorithm was originally introduced in \cite{dsnn_conversion_ijcnn} and extended in \cite{dsnn_conversion_abhronilfin} to improve accuracy on deep models. It is applicable to IF neuron models only. The algorithms works as follows.   
First, a constrained ANN model with ReLU activation but no bias, max-pool, or BN layers is trained. Second, the trained ANN weights are copied to an SNN model and the analog input training data of the ANN is converted into rate-coded input spikes through a Poisson event generation process over $T$ time steps (detailed in Section \ref{subsubsec:input}). Finally, the firing threshold $v_{th}$ of each layer is set to the maximum value of the $\sum w_{ij}O_j^t$ (the $2^{nd}$ term in Eq. \ref{eq:neuron_discrete}) evaluated over the $T$ time steps and computed by using a subset of the training images.  This threshold tuning operation ensures that the IF neuron activity  mimics precisely the ReLU function of the corresponding ANN. 
Even though this conversion technique largely mitigates the training complexity of deep SNNs and achieves state-of-the-art inference, the resulting SNNs generally have large inference latency ($T \approx 2,500$),
which limits their energy efficiency.
For our SNN models we adopt a hybrid training strategy \cite{rathi2020iclr} where we use the SNN training for only a few epochs by using a linear surrogate-gradient \cite{bellec_2018long} based SL, as will be detailed in Section \ref{sec:snn_agc}.  

\subsubsection{Hybrid SNN Training}
\label{subsec:snn_hybrid}
Recently reference \cite{rathi2020iclr} proposed a hybrid training methodology where ANN-SNN conversion is performed as an initialization step, which is followed by running the approximate gradient descent algorithm on the initialized network. The authors claimed that combining the two training techniques helps the SNN converge within a few epochs while requiring fewer time steps. Although the authors demonstrated competitive classification accuracy on complex datasets, the fine-tuning step with spike-based backpropagation 
over multiple time steps remains compute-expensive.

\subsection{Model Compression of SNNs}
\label{subsec:modelcompress}
To improve the energy-efficiency of the SNN models, reference \cite{soft_pruning} developed a pruning methodology that uses the sparse firing characteristics of IF neurons to adjust the corresponding number of weight updates during training.
Other works use dynamic pruning either by leveraging multi-strength SNN (M-SNN) models \cite{dynamic_prune} or the correlation  
between pre and post-neuron spike activities \cite{rathi2017stdp}.
However, most of these works evaluated their approaches on shallow architectures for small datasets such as MNIST and DVS. 
Recently, reference \cite{deng2019comprehensive} used ADMM to compress the models while performing STBP-based SNN training. However, as mentioned earlier, ADMM requires additional hyperparameters of per-layer target parameter density, which in turn necessitates iterative training \cite{liu2020autocompress} or complex procedures such as reinforcement learning. This makes the already-tedious SNN training more difficult. Moreover, these method fail to provide conventional (ANN-equivalent) accuracy metrics for the compressed models. 
Recently, some researchers have also focused on model quantization \cite{quantize_snn2, lu2020exploring}. 
We point out that although our proposed approach focuses on pruning, it can easily be extended to support pruning on quantized models, as these are largely orthogonal techniques.
  
To address these challenges, we propose to prune the constrained ANN models (designed for SNN conversion) using 
attention-guided compression.
This strategy, which is detailed in Section \ref{subsec:ann_agc}, requires only a global target parameter density and performs sparse weight updates (sparse-learning) to avoid iterative training. The specific sparse-learning we adopt \cite{dettmers2019sparse} is computationally more efficient than other strategies \cite{bellec2017deep} and uses a more comprehensive approach of regrowing the weights based on the magnitude of their momentum, outperforming similar approaches \cite{bellec2017deep, mostafa2019parameter}. Our hypothesis is that our proposed approach can target ANNs designed for SNN conversion, accelerate the current tedious training procedures for SNN compression, and yield superior inference accuracy.



\section{Proposed SNN Training Method}
\label{sec:snn_agc}

This section describes our two-step hybrid sparse-learning strategy.
Section \ref{subsec:ann_agc} describes our AGC training framework that targets conversion-friendly compressed ANNs while Section \ref{subsec:channel_prune} details the extensions that support channel pruning. 
Section \ref{subsec:snn_compress} then presents our SL based supervised SNN training. 

\vspace{-2.5mm}
\subsection{ANN Training with AGC}
\label{subsec:ann_agc}

An activation-based mapping $\mathcal{F}$ can convert a 3D tensor of activations of a convolutional layer $l$, $\textbf{A}^l \in \mathbb{R}^{H_{i}^l \times W_{i}^l \times C_{i}^l}$ with $H_i^l \times W_i^l$ feature planes and $C_i^l$ channels to a 2D attention map, i.e.,
 \begin{align}
     \mathcal{F}: \mathbb{R}^{H_{i}^l \times W_{i} \times C_{i}} \rightarrow \mathbb{R}^{H_i \times W_i}
     \label{eq:atten_map}
 \end{align}
{A commonly used attention map function is, $\mathcal{F}^p$ = $\sum_{c=1}^{C_i^l}|\textbf{A}^l(:,:,c)|^p$, i.e. the sum of absolute values of layer activation values across all the channels raised to the power of $p$. Here $p \geq 1$ is a parameter choice to determine the relative degree of emphasis the most discriminative parts of the feature map should have.} 
These activation maps coalesce important features of a model's intermediate layers and are typically used for  distillation-based training from a compute-intensive teacher model to a less complex student model \cite{zagoruyko2016paying}. 

Inspired by the success of attention transfer in conventional distillation, we introduce a meta-model $\Psi_m$ to guide the model compression of a BN-free ANN model $\Psi_c$. 
Our goal is to make the activation maps of the compressed model closely follow (in the sense which is defined below) those of an unpruned meta model to mitigate the gradient-explosion in the initial steps of training of the compressed model. Let $x_i$ and $y_i$ denote the $i^{th}$ input sample and the corresponding ground truth output of dataset $\mathcal{X}$. Let $p_{f^{{\Psi_c}}}(x_i)$ represent the logit response of $x_i$ for ${\Psi_c}$. Moreover, let $Q^{\Psi_c}_j$ and $Q^{\Psi_m}_j$ denote the $j^{th}$ pair of vectorized versions of attention maps $\mathcal{F}$ of specific layers of $\Psi_c$ and $\Psi_m$, respectively. Then, the proposed training loss function may be defined as, 
\begin{align}
    {\mathcal{L}} &= 
     \frac{\alpha}{2}\sum_{j \in \mathcal{I}}\left\lVert\frac{Q^{\Psi_c}_j}{\lVert Q^{\Psi_c}_j\rVert_2} - \frac{Q^{\Psi_m}_j}{\lVert Q^{\Psi_m}_j\rVert_2}\right\lVert_2 \\\nonumber
     &+ \sum_{(x_i, y_i) \in \mathcal{X}}{\mathcal{X}}{\mathcal{E}}(\sigma(p_{f^{\Psi_c}}(x_i)), y_i) ,
\label{eq:agc_loss}
\end{align}
\noindent

In Eq. 5 the loss function has two major components. The first component is the activation-based scaled attention-guided loss (AG-loss) term, which aims to minimize the differences between $\Psi_m$ and $\Psi_c$'s activation maps where $\alpha$ denotes a scale factor. 
The second component is the standard cross-entropy loss (CE-loss) of $\Psi_c$.
In our case, the $\Psi_m$ is either a low-complexity unpruned model or an unpruned variant of the $\Psi_c$, in contrast to the compute-intensive teacher models used in distillation. 
Because the purpose of $\Psi_m$ is to avoid gradient explosion during the initial steps of training, we remove the attention-guided (AG) loss component by setting $\alpha$ to zero after $\epsilon$ epochs. This allows $\Psi_c$ to not be upper-bounded by the performance of $\Psi_m$.\footnote{We note that some distillation approaches also penalize the network using a weighted KL-divergence between the probabilistic outputs of the two networks particularly for a more complex teacher model. However, we empirically verified that adding KL-divergence to the loss worsens performance of $\Psi_c$. Also, this added term reduces the importance of the ${\mathcal{L}}^{\Psi_c}_{CE}$ which is critical in sparse-learning.
}  
To evaluate the AG-loss component, we compute the $l_2$-norm of the difference between the $l_2$-normalized attention-maps and accumulate these differences over all layer pairs in $j \in \mathcal{I}$.  Although pairs of layers that have different shapes can be supported, we generally choose pairs of layers where the spatial dimensions of $\Psi_m$ and $\Psi_c$ are similar. The fact that the $\Psi_m$ can be any low complexity model, as opposed to $\Psi_c$, reduces the computational complexity of pre-training and increases the pre-trained model re-usability compared to  distillation.

To satisfy the non-zero parameter budget associated with a user-given target parameter density $d$,
we start ANN training with initialized weights and a random binary prune-mask meeting the budget set by $d$. We then compute the normalized momentum of the non-zero weights (evaluated using the loss function of Eq. \ref{eq:agc_loss}) during an epoch and dynamically evaluate the layer significance for the given density $d$. After pruning a fixed number of least-magnitude weights from each layer, based on this momentum knowledge, we update the pruning mask to re-grow weights in the important layers 
\cite{dettmers2019sparse}.  Details of the AGC framework are presented in Algorithm \ref{alg:agc}. 

\textbf{Observation 1:} \textit{AGC improves sparse learning of BN-free models because activation maps of the sparse model are designed so as to follow that of an uncompressed model.}

We know BN-free models are encouraged to suppress the range of activation during the initial phase of the training \cite{shao2020normalization} to avoid gradient explosion. Our empirical results showing successful convergence of AGC-based learning demonstrate that we can restrict the activation map values via our proposed loss in Eq. 5. We exemplify this in Fig. \ref{fig:agc_framework}(c) showing the successful compression of ResNet12 to a target density $d=0.1$ using our proposed AGC framework, contrasting it with the significant accuracy drop observed when compressed using SL \cite{dettmers2019sparse}. 

\begin{algorithm}[t]
\footnotesize
\SetAlgoLined
\DontPrintSemicolon
\textbf{Input}: \text{runEpochs},  momentum $\pmb{\mu}^l$, prune rate $p$, initial $\textbf{W}$, initial $\textbf{M}$, prune type $p_{type}$, target density $d$, $\Psi_m$, $\epsilon$.\\
\KwData{$ i = 0..\text{runEpochs}$, pruning rate $p = p_{i=0}$}\;
\For{$l \leftarrow 0$ \KwTo $L$}
{
    ${\textbf{W}}^l \leftarrow \text{init}({\textbf{W}}^l)$\ \&
    $\textbf{M}^l \leftarrow \text{createMaskForWeight}({\textbf{W}}^l, d)$\;  
    $\text{applyMaskToWeights}({\textbf{W}}^l, \textbf{M}^l)$\;
}
\For{$\text{i} \leftarrow 0$ \KwTo \text{runEpochs}}
{
    $\alpha = \alpha *Bool(i < \epsilon)$\;
    \For{$\text{j} \leftarrow 0$ \KwTo \text{numBatches}}
    {
     ${\mathcal{L}} = \alpha*{\mathcal{L}}_{AG} + {\mathcal{L}}_{CE}$\;
     $\frac{\partial {\mathcal{L}}}{\partial{\textbf{W}}} = \text{computeGradients(}{\textbf{W}}, {\mathcal{L}}\text{)}$\;
     $\text{updateMomentumAndWeights}(\frac{\partial {{\mathcal{L}}}}{\partial{\textbf{W}}}, \pmb{\mu} )$\;
     \For{$l \leftarrow 0$ \KwTo $L$}
     {
     $\text{applyMaskToWeights}({\textbf{W}}^l, \textbf{M}^l)$\;
     }
    }
    $\text{tM} \leftarrow \text{getTotalMomentum}({\pmb{\mu}})$\;
    $\text{pT} \leftarrow \text{getTotalPrunedWeights}({\textbf{W}}, p_i)$\;
    $p_i \leftarrow \text{linearDecay}(p_i)$\;
    \For{$l \leftarrow 0$ \KwTo $L$}
    {
        $\pmb{\mu}^l \leftarrow \text{getMomentumContribution}({\textbf{W}}^l,
        \textbf{M}^l, \text{tM}, \text{pT}, p_{type})$\;
        $\text{Prune}({\textbf{W}}^l, \textbf{M}^l, p_i, \text{pT}, p_{type})$\;
  $\text{Regrow}({\textbf{W}}^l, \textbf{M}^l, \pmb{\mu}^l \cdot \text{tM}, \text{pT}, p_{type})$\;
  $\text{applyMaskToWeights}({\textbf{W}}^l, \textbf{M}^l)$\;
  }
}
 \caption{Detailed algorithm for attention guided compression.}
 \label{alg:agc}
\end{algorithm}

\subsection{Support for Structured Channel Pruning}
\label{subsec:channel_prune}

Let the weight tensor of a convolutional layer $l$ be denoted as $\textbf{W}^l \in \mathbb{R}^{C^l_o \times C^l_i \times k^l \times k^l}$, where $k^l$, $C^l_o$, and $C^l_i$ represents the height (and width) of the convolutional filter, the number of filters, and the number of channels per filter, respectively. We convert this tensor to a 2D weight matrix, with $C^l_o$ rows and $C^l_i \times k^l \times k^l$ columns. 
We then partition this matrix into $C^l_i$ sub-matrices of $C^l_o$ rows and $k^l \times k^l$ columns, one for each channel. We rank the importance of a channel $c$ by evaluating the Frobenius norm (F-norm) of the corresponding sub-matrix, effectively computing $F^l_c$ = $\lvert{\textbf{{W}}^l_{:,c,:,:}}\rvert^{2}_F$. 
Based on the fraction of non-zero weights that need to be rewired during an epoch $i$, set by the pruning rate $p_i$, we compute the number of channels that must be pruned from each layer, ${c}_{p_i}^l$, and prune the ${c}_{p_i}^l$ channels with the lowest F-norms. 
We then compute each layer's importance based on the normalized momentum contributed by its non-zero channels, and use it to determine the number of zero-F-norm channels $r^l_i \geq 0$ that should be re-grown in that layer. 
In particular, we re-grow the $r^l_i$ zero-F-norm channels with the highest F-norms of their momentum.
It is noteworthy that for structured pruning we only prune the convolutional layers because the FC-layers do not have a notion of channels, whereas for irregular pruning we prune both the CONV and FC layers.

\vspace{-1mm}
\subsection{Direct Input Encoded SNN Training via Sparse-learning}
\label{subsec:snn_compress}

After compressing the ANN model, we compute the threshold of each layer via the threshold evaluation strategy presented in \cite{dsnn_conversion_abhronilfin}. We then perform SNN training for a few epochs to reduce the number of inference time steps. The standard supervised SNN training uses the surrogate gradient method \cite{bellec_2018long, wu_backprop, zenke2018superspike} to facilitate backpropagation-based optimization for non-differentiable neuron spikes. 
The surrogate gradient is typically a pseudo-derivative that approximates the gradient as a linear or an exponential function of the membrane potential. 
Also, similar to \cite{rueckauer2017conversion}, the pixel intensities of an image are applied directly to the SNN input layer at each time step. 

Existing SNN training schemes must be adjusted in the context of our proposed compression framework. 
In particular, in our SNN training the neuron membrane dynamics are modeled as
\vspace{-1.5mm}
\begin{align}
    u_i^{t+1} &= \lambda{u_i^t} + \sum_j m_{ij}*w_{ij}*O_j^t - v_{th}O_i^t\\
   O_i^{t}   &=
    \begin{cases}
    1, & \text{if } z_i^{t}>0, \\
    0, & \text{otherwise}
    \end{cases} 
    \label{eq:sparse_snn_train}
\end{align}
\noindent
where $z_i^{t}$ = ($\frac{u_i^t}{v_{th}} -1$) denotes the normalized membrane potential and $m_{ij} \in \{0,1\}$ is the fixed prune mask of a neuron $i$ and its pre-synaptic neuron $j$, 
where a 0 and 1 indicate absence and presence of synaptic weights, respectively. Note that $O_i^t$ is the $i^{th}$ post synaptic neuron activation at time $t$. 
The first convolutional layer extracts the input features and produces spikes following the LIF model by using multiply-accumulate operations (involving the non-binary $O_j^t$ values corresponding to input pixels) and generate output spikes whenever the accumulated value crosses the layer threshold. The subsequent layers, in contrast, generate output spikes according to accumulated weighted binary input spikes, a process that does not involve any multiplications.
In the last layer we accumulate the inputs over all time steps and compute their softmax as a model of their multi-class probability. In addition to adopting the direct encoding scheme, our proposed training method also learns the weights, membrane leak, and firing threshold parameters for each layer.
Such a joint optimization procedure enables the SNN to achieve high accuracy with low inference latency. Interestingly, tuning both the threshold and leak is consistent with biological processes as previous research has shown that neurons in different cortical regions of the mammalian brain operate with different leak \cite{tunable_leak} and threshold \cite{tunable_threshold} values.

During backpropagation with a learning rate $\eta$, the linear layer weights are updated as
\begin{align}
        & w_{ij} = w_{ij} - \eta\delta w_{ij} \\
    & \delta w_{ij} = m_{ij} * \sum_t \frac{\partial\mathcal{L}}{\partial O_i^t}\frac{\partial O_i^{t}}{\partial z_i^{t}}\frac{\partial z_i^{t}}{\partial u_i^{t}}\frac{\partial u_i^{t}}{\partial w_{ij}^t} 
    \label{eq:sparse_grad_update}
\end{align}

\noindent
where the term $\frac{\partial O_i^{t}}{\partial z_i^{t}}$ requires a pseudo-derivative. We follow \cite{bellec_2018long} to define this as
\begin{align}
    \frac{\partial O_i^{t}}{\partial z_i^{t}} = \gamma * max\{0, 1-|z_i^{t}|\} 
    \label{eq:linear_surrogate}
\end{align}
\noindent
where $\gamma$ is a damping factor for the backpropagation error.
%
The threshold update is computed as
\begin{align}
   v_{th}&=v_{th}-\eta\delta{v_{th}} \\
   \delta{v_{th}}&=\sum_{t}\frac{\partial\mathcal{L}}{\partial O_i^t}\frac{\partial O_i^t}{\partial z_i^t}\frac{\partial z_i^t}{dv_{th}} 
\end{align}

\noindent
Finally, the leak term is updated as
\begin{align}
   \lambda&=\lambda-\eta\delta{\lambda} \\
   \delta{\lambda}&=\sum_{t}\frac{\partial\mathcal{L}}{\partial O_i^t}\frac{\partial O_i^t}{\partial z_i^t}\frac{\partial z_i^t}{\partial u_i^t}\frac{\partial u_i^t}{\partial \lambda}
\end{align}

As we update only weights with a corresponding mask of 1, we refer to this as `sparse-learning' and the entire approach as a form of hybrid SL. 

\section{Experiments}
\label{sec:expt}

This section first details the experimental setup used to perform our training.
We then evaluate the effectiveness of the proposed compression scheme by presenting results on CIFAR-10, CIFAR-100, and Tiny-ImageNet with VGG and ResNet model variants.
Finally, we demonstrate the compute energy-efficiency of the generated hybrid MAC-AC based models over the uncompressed baselines.

\subsection{Experimental Setup}
\label{subsec:setup}
\subsubsection{Input Data}
\label{subsubsec:input}
\begin{figure}[!t]
\includegraphics[width=0.49\textwidth]{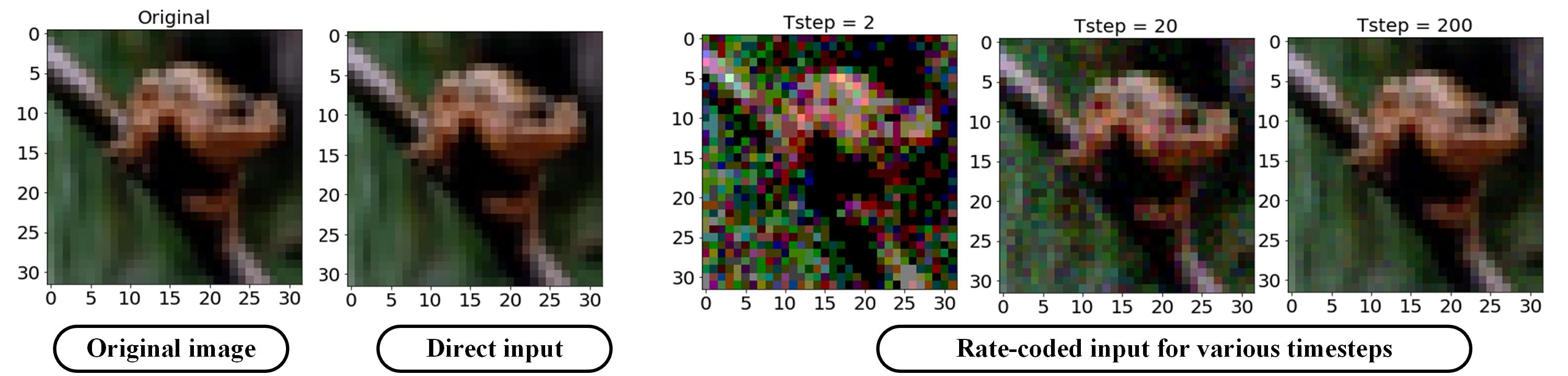}
\centering
   \caption{Original input along with its  direct-coded input and  rate-coded spike equivalent input images for different number 
   of time steps $T$.}
\label{fig:input_rate_coded_image}
\vspace{-2mm}
\end{figure}

For both AGC based ANN and direct input coded SNN training, we use the standard data-augmented (horizontal flip and random crop with reflective padding) input. For rate-coded input based SNN training, we use a Poisson generator function to produce a spike train with rate proportional to the input pixel value as input data for the ANN-to-SNN conversion and SNN training \cite{Kundu2021WACV}. Fig. \ref{fig:input_rate_coded_image} shows both the direct input and rate coded inputs corresponding to an original image. Note that, as $T$ increases, the rate-coded input spike train becomes a closer approximation to the direct input.

\begin{table}[!t]
\begin{center}
\scriptsize\addtolength{\tabcolsep}{-4.5pt}
{\makegapedcells
\begin{tabular}{|c|c|c|c||c|c|c|}
\hline
 Dataset  & \multicolumn{3}{|c|}{$T_d$} & \multicolumn{3}{|c|}{$T_r$} \\
 \cline{2-7}
          & VGG11 & VGG16 & ResNet12 & VGG11 & VGG16 & ResNet12\\
\hline
CIFAR10  & 10 &  10  & 25 & 100 & 100 & 175 \\
CIFAR100 & 10 &  10  & 25 & 120 & 120 & 200 \\
Tiny-ImageNet & -- &  10  & -- & -- & 150 & -- \\
\hline
\end{tabular}}
\end{center}
\caption{Different values of reduced time steps $T_d$ and $T_r$ for models on different datasets.}
\label{tab:tstep_values}
\vspace{-4mm}
\end{table}

\begin{table*}[!t]
\begin{center}
\scriptsize\addtolength{\tabcolsep}{-4.5pt}
{\makegapedcells
\begin{tabular}{|c|c|c|c|c|c|c||c|c|c|c|c|}
\hline
 & Compre- & Channel & $a$. & \multicolumn{2}{|c|}{$b1$. Accuracy ($\%$) with} & $c1$. Accuracy ($\%$) & \multicolumn{2}{|c|}{$b2$. Accuracy ($\%$) with} & $c2$. Accuracy ($\%$) & Accuracy & Latency   \\
 
 Architecture & ssion & Compre & ANN ($\%$) &\multicolumn{2}{|c|}{ANN-to-SNN conversion} & after sparse & \multicolumn{2}{|c|}{ANN-to-SNN conversion} & after sparse & Improvement & Improvement \\
 
 & ratio & -ssion  & accuracy &  \multicolumn{2}{|c|}{(direct input)}& SNN training & \multicolumn{2}{|c|}{(rate coded input)} & SNN training & ($c1-c2$) & $\frac{T_r}{T_d}$ \\
\cline{5-6}\cline{8-9}
 &   &  ratio &   &  $T$ = 100  &  $T_d$    &   &  $T$ = 2500  &  $T_r$  &  &  &  \\ 
\hline
\hline
 \multicolumn{12}{|c|}{Dataset : CIFAR-10} \\
\hline
\hline
 VGG11  & $1\times$  & $1 \times$ & 91.57 & 88.02 & 75.49 &  90.37 & 91.17 & 89.16 & 89.84 & +0.53 & $10\times$ \\
        & $10\times$ & $1 \times$ & 91.10 & 83.76 & 26.97 & 90.94 & 90.64 & 86.16 & 90.45 & +0.49 & \\
\hline
 VGG16 & $1 \times$ & $1 \times$  & 92.55  & 90.66  & 77.81   &  91.52 & 92.01  &  84.79  &  91.13 & +0.39 & $10\times$ \\
        & $2.5 \times$ & $1 \times$ & 92.97  & 91.37 & 85.82  & \textbf{92.53} & 92.92 & 90.08  & 91.29 & +1.24 &  \\
        & $20 \times$  & $1.02 \times$ & 91.85  & 85.90 & 84.71  &  91.09 & 91.39 & 79.08  &  90.74 & +0.35 & \\
        & $\textbf{33.4} \times$ & $1.15 \times$ & 91.79 & 83.85 & 82.37  &  90.59 & 91.22 & 72.53  &  90.15 & +0.44  &  \\
\hline
 ResNet12 & $1 \times$ & $1 \times$ & 91.37 & 87.35  & 85.68 & 90.88 & 90.87  & 88.98 & 90.41 & +0.47 &  $7\times$   \\
        & $10 \times$ & $1.20 \times$ & 92.04 & 89.70 & 84.41 & 91.08 & 91.71 &  83.46  & 90.79 & +0.29 &   \\
\hline
\hline
 \multicolumn{12}{|c|}{Dataset : CIFAR-100}  \\
\hline
\hline
 VGG11  & $1\times$  & $1 \times$ & 66.30 & 56.40 & 43.78 & 62.03 & 64.18 & 62.49 & 64.37 & -2.34 & $12\times$ \\
        & $4\times$ & $1 \times$ & 67.40 & 59.74 & 41.72 & \textbf{65.34} & 65.10 & 62.57 & 64.98 & +0.36 &  \\
\hline
 VGG16  & $1\times$  & $1 \times$ & 67.62 & 58.74 & 41.61 & 64.8 & 65.91 & 54.30 & 64.69 & +0.11 & $12\times$   \\
        & $\textbf{10}\times$ & $1.03 \times$ & 67.45 & 57.99 & 34.2 & 65.32 & 65.84 & 51.63 & 64.63 & +0.69 &  \\
\hline
 ResNet12  & $1\times$  & $1 \times$ & 61.61 & 59.00 & 49.87 & 60.46 & 59.85 & 56.97 & 62.60 & -2.14 & $8\times$  \\
        & $10\times$ & $1.31 \times$ & 63.52 & 49.89 & 34.09 & 62.28 & 61.43 & 52.66 & 63.02 & -0.74 &   \\
\hline
\hline
 \multicolumn{12}{|c|}{Dataset : Tiny-ImageNet} \\
\hline
\hline
 VGG16  & $1\times$  & $1 \times$ & 56.56 & 45.75 & 22.87 & 53.37 & 56.80 & 51.14 & 51.92 & +1.45 & $15\times$ \\
        & $\textbf{2.5}\times$ & $1 \times$ & 57.00 & 46.13 & 19.94 & \textbf{54.10} & 56.06 & 51.90 & {52.70} & +1.40 &  \\

\hline
\end{tabular}}
\end{center}
\caption{Direct and rate coded input based model performances with AGC based irregular compression on CIFAR-10, CIFAR-100, and Tiny-ImageNet after ANN training ($a$), ANN-to-SNN conversion ($b1$/$b2$), and SNN training ($c1$/$c2$).}
\label{tab:snn_prune_results_direct_input}
\vspace{-5mm}
\end{table*}

\subsubsection{Model and ANN Training}
\label{subsubsec:model_and_ann}
To facilitate near loss-less SNN conversion we adopted various recommended constraints for the ANN models including no use of BN and maxpool layers \cite{dsnn_conversion_abhronilfin}. For the uncompressed models we used dropout in both CONV and linear layers as a regularizer. However, we reduce the CONV layer dropout values for the compressed ANNs because they have lower chance of over fitting  \cite{Kundu2021WACV}. For our ResNet models, instead of the initial CONV layer, we use a pre-processing block consisting of three convolution layers of size $3\times 3$ with a stride of 1.

We performed ANN training for 240 epochs with an initial learning rate (LR) of $0.01$ that decayed by a factor of 0.1 after $150$, $180$, and $210$ epochs. We hand tuned and set both $\alpha$  and $\epsilon$ epoch to be $100$. We chose to linearly decay the prune rate every epoch that starts with a value of 0.5.  Unless stated otherwise, for the meta-model we used an unpruned VGG9 ANN designed and trained with the same constraints.

\subsubsection{Conversion and SNN Training}

We first examine the distribution of the neuron input values over the total $T$ time steps across all neurons of the first layer for a small batch of input images (of size 512 in our case) and set the layer threshold to 99.7 percentile of the scaled value of the evaluated threshold \cite{rathi2020diet}. In our experiments we scale the initial thresholds by 0.4.
Similarly, we then compute the thresholds of the subsequent layers sequentially by examining the distribution of their input 
values.\footnote{For the ResNet variant, we evaluated thresholds only for the pre-processing block convolution layers \cite{rathi2020iclr}.} We keep the leak of each layer set to unity while evaluating initial thresholds. At the start of SNN training, we initialize the weights with those from the trained ANN and initialize the leak parameters to $1.0$. 
We then perform the sparse-learning based SNN training for only 20 epochs. We set $\gamma$ = $0.3$ \cite{bellec_2018long} and for direct input based training used the ADAM optimizer with a starting LR of $10^{-4}$ which decays by a factor of $0.5$ after 12, 16, and 18 epochs. Further SNN training details for rate coded inputs are described in \cite{Kundu2021WACV}.

\begin{figure}[!t]
\includegraphics[width=0.40\textwidth]{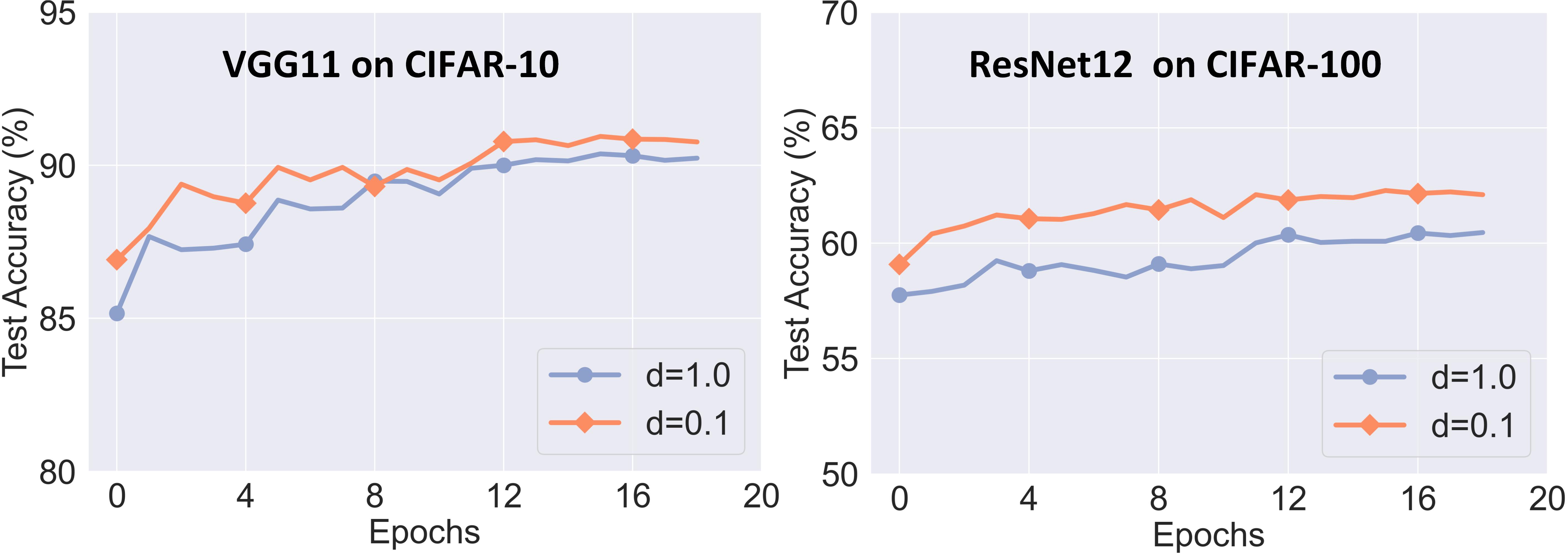}
\centering
   \caption{Test accuracy vs. epochs for two different models on two datasets.}
\label{fig:test_acc_snn_plots}
\vspace{-3mm}
\end{figure}

\subsection{Compression Results}
\label{subsec:compression_res}
Table \ref{tab:snn_prune_results_direct_input} shows the performance of irregularly compressed models obtained using our proposed compression scheme for all three datasets on both direct and rate coded inputs. We denote as $T_d$ and $T_r$ the number of time steps for direct and rate coded inputs, respectively. In particular, to have competitive test accuracy we choose different values of these two variables based on models and datasets, as detailed in Table \ref{tab:tstep_values}. As shown in Table \ref{tab:snn_prune_results_direct_input}, the direct input coded models not only require up to $15\times$ lower inference latency, but they also have higher accuracy in most cases. Here, the compression ratio is computed as the inverse of target density. The results show that sparse-learning based SNN training significantly improves model performance with reduced time steps and direct input coding. In particular, for VGG11 on CIFAR-10 with a compression ratio of $10\times$, our hybrid SL improves inference accuracy by $\mathord{\sim}64\%$ when compared to the original conversion-based models with the same reduced $T_d$.  With a compression ratio of up to $33.4\times$, $10\times$, and $2.5\times$ for CIFAR-10, CIFAR-100, and Tiny-ImageNet, respectively, our hybrid SL trained VGG models perform similar to their uncompressed counterparts requiring only 10 times steps. We hypothesize that for lower compression ratios these improvements in accuracy may be because AGC acts as an effective regularizer. 
For example, the VGG16 model on CIFAR-10 (direct input) with $2.5\times$ compression ratio improves classification performance by $1.01\%$ compared to the uncompressed baseline which only uses dropout for regularization. 

Table \ref{tab:snn_channel_prune_results_direct_input} shows the model performance with channel pruning on both types of input coding technique. Similar to irregularly pruned models, we observe the channel pruned models with direct input outperforms those with rate coded input both in terms of inference latency and model performance. 

\textbf{Observation 2:} \textit{Compressed SNN models shows faster convergence trend compared to uncompressed counter part.}

As we can observe both from the model performance in Table \ref{tab:snn_prune_results_direct_input} and the accuracy vs epochs plots in Fig. \ref{fig:test_acc_snn_plots}, the compressed SNN models show improved classification accuracy with faster convergence. This may be because an ANN sparsely trained via AGC with target density $d < 1.0$ acts as a well regularized model.
However, models with extremely small densities $d$ incur small accuracy drops likely because of the limited learnability of the model.

\textbf{Observation 3:} \textit{Structured SNN pruning yields reduced compression compared to irregular pruning for similar accuracy performance, yet perform considerably better in effective channel-size reduction.}

As we can see in Table \ref{tab:snn_prune_results_direct_input} and \ref{tab:snn_channel_prune_results_direct_input}, channel pruning provides reduced compression of around an order of magnitude with little to no drop in accuracy. However, channel compression potentially speeds up inference because of the reduced number of channels. In particular, for VGG16 with only $5\times$ compression, channel pruning provides a channel compression\footnote{Channel compression ratio is defined to be the ratio of total number of channels to number of non-zero F-norm channels of a model.} of $3.55\times$ compared to only $1.15\times$ with irregular pruned model of target compression of $33.4\times$.

\begin{table*}[!t]
\begin{center}
\scriptsize\addtolength{\tabcolsep}{-4.5pt}
{\makegapedcells
\begin{tabular}{|c|c|c|c|c|c|c||c|c|c|c|c|}
\hline
 & Compre- & Channel & $a$. & \multicolumn{2}{|c|}{$b1$. Accuracy ($\%$) with} & $c1$. Accuracy ($\%$) & \multicolumn{2}{|c|}{$b2$. Accuracy ($\%$) with} & $c2$. Accuracy ($\%$) & Accuracy & Latency   \\
 
 Architecture & ssion & Compre & ANN ($\%$) &\multicolumn{2}{|c|}{ANN-to-SNN conversion} & after sparse & \multicolumn{2}{|c|}{ANN-to-SNN conversion} & after sparse & Improvement & Improvement \\
 
 & ratio & -ssion  & accuracy &  \multicolumn{2}{|c|}{(direct input)}& SNN training & \multicolumn{2}{|c|}{(rate coded input)} & SNN training & ($c1-c2$) & $\frac{T_r}{T_d}$ \\
\cline{5-6}\cline{8-9}
 &   &  ratio &   &  $T$ = 100  &  $T_d$    &   &  $T$ = 2500  &  $T_r$  &  &  &  \\ 
\hline
\hline
 \multicolumn{12}{|c|}{Dataset : CIFAR-10} \\
\hline
\hline
 VGG11  & $2\times$ & $1.91 \times$ & 91.89 & 89.89 & 82.86 & 90.97 & 91.43 & 89.30 & 91.00 & -0.03 & $10 \times$\\
\hline
 VGG16       & $5 \times$  & $3.55 \times$   & 91.24  & 84.89  & 75.01   &  90.54 & 90.86 & 84.49 & 89.99 & +0.55 & $10 \times$ \\
\hline
 ResNet12 & $2 \times$ & $1.37 \times$ & 92.91 & 89.23  & 83.80 & \textbf{92.44} & 92.60 & 37.73 & 91.81 & +0.63 & $7 \times$ \\
\hline
\hline
 \multicolumn{12}{|c|}{Dataset : CIFAR-100} \\
\hline
\hline
 VGG11  & $2\times$  & $1.91 \times$ & 65.27 & 56.67 & 37.78 & 60.38 & 63.59 & 60.46 & 63.67 & -3.29 & $12 \times$ \\
\hline
 VGG16  & $2.5\times$  & $2.06 \times$ & 68.62 & 61.22 & 43.19 & \textbf{65.78} & 67.38 & 49.94 & 65.9 & -0.12 & $12 \times$ \\
\hline
 ResNet12  & $2\times$  & $1.36 \times$ & 63.82 & 54.88 & 54.43 & 62.82 & 62.53 & 56.25 & 63.18 & -0.36 & $8 \times$\\
\hline
\hline
 \multicolumn{12}{|c|}{Dataset : Tiny-ImageNet} \\
\hline
\hline
 VGG16  & $2.5\times$  & $2.06 \times$ & 54.77 & 39.47 & 22.58 & 50.53 & 50.00 & 41.50 & 49.46 & +1.07 & $15 \times$\\
\hline
\end{tabular}}
\end{center}
\caption{Direct and rate coded input based model performances with AGC based structured channel compression on CIFAR-10, CIFAR-100, and Tiny-ImageNet after ANN training ($a$), ANN-to-SNN conversion ($b1$/$b2$), and SNN training ($c1$/$c2$).}
\label{tab:snn_channel_prune_results_direct_input}
\vspace{-3mm}
\end{table*}

\begin{table}
\begin{center}
\scriptsize\addtolength{\tabcolsep}{-3.5pt}
{
\begin{tabular}{|c|c|c|c|c|c|}
\hline
Authors &  Training & Architecture & Compress- & Accuracy & Time  \\
 & type &  & ion ratio & ($\%$) & steps \\
\hline
\hline
 \multicolumn{6}{|c|}{Dataset : CIFAR-10} \\
\hline
\hline
Cao et al. & ANN-SNN & 3 CONV, & $1 \times$ & 77.43 & 400 \\
(2015) \cite{cao2015spiking} & conversion & 2 linear &  & & \\
\hline
Sengupta et  & ANN-SNN & VGG16 & $1 \times$ & 91.55 & 2500 \\
al. (2019)\cite{dsnn_conversion_abhronilfin} & conversion &  &  & & \\
\hline
Wu et al. & Surrogate & 5 CONV, & $1 \times$ & 90.53 & 12 \\
(2019) \cite{wu2019direct} & gradient & 2 linear &  & & \\
\hline
Rathi et al. & Hybrid & VGG16 & $1 \times$ & 91.13 & 100 \\
(2020) \cite{rathi2020iclr} & training  & & $1 \times$  & {92.02} & 200 \\
\hline
Rathi et al. & DIET & VGG16 & $1 \times$ & 91.25  & \textbf{5} \\
(2020) \cite{rathi2020diet} & SNN  & &  & &  \\
\hline
Deng et al. & STBP & 11 layer & $1 \times$ & 89.53 & {8} \\
(2020) \cite{deng2019comprehensive}& training & CNN &  & & \\
\hline
Deng et al. & STBP & 11 layer & $4 \times$ & 87.38 & 8 \\
(2020) \cite{deng2019comprehensive}& training & CNN &  & & \\
\hline
\rowcolor{Gray}
This work & Hybrid SL& VGG16 & $2.5 \times$ & {91.29} & 100 \\
\rowcolor{Gray}
 & (rate-coded) &  & $\textbf{33.4} \times$ & 90.15 & 100\\
 \hline
 \rowcolor{Gray}
 This work & Hybrid SL & VGG16 & $2 \times$ & \textbf{92.74} & 10 \\
\rowcolor{Gray}
 & (direct) &  & $\textbf{33.4} \times$ & 90.59 & 10\\
\hline
\hline
 \multicolumn{6}{|c|}{Dataset : CIFAR-100} \\
\hline
\hline
Rathi et al. & DIET & VGG16 & $1 \times$ & 64.89 & \textbf{5} \\
(2020) \cite{rathi2020diet} & SNN  & &  & &  \\
\hline
Deng et al. & STBP & 11 layer & $2 \times$ & 57.83 & 8 \\
(2020) \cite{deng2019comprehensive}& training & CNN &  & & \\
\hline
\rowcolor{Gray}
This work & Hybrid SL & VGG11 & $ {4} \times$ & {64.98} & 120 \\
\rowcolor{Gray}
 & (rate-coded) &  &  &  &  \\
\hline
\rowcolor{Gray}
This work & Hybrid SL & VGG11 & $\textbf{4} \times$ & \textbf{65.34} & 10 \\
\rowcolor{Gray}
 & (direct) &  &  &  &  \\
\hline
\end{tabular}}
\end{center}
\caption{Performance comparison of the proposed hybrid SL with state-of-the-art deep SNNs on CIFAR-10 and CIFAR-100. Note for \cite{rathi2020diet} the original paper reported accuracy of after SNN training of 100 epochs. However, here we choose to provide best test accuracy after 20 epochs from the log file of the authors' shared codebase.}
\label{tab:snn_comparison}
\vspace{-5mm}
\end{table}

\begin{figure}[!t]
\includegraphics[width=0.40\textwidth]{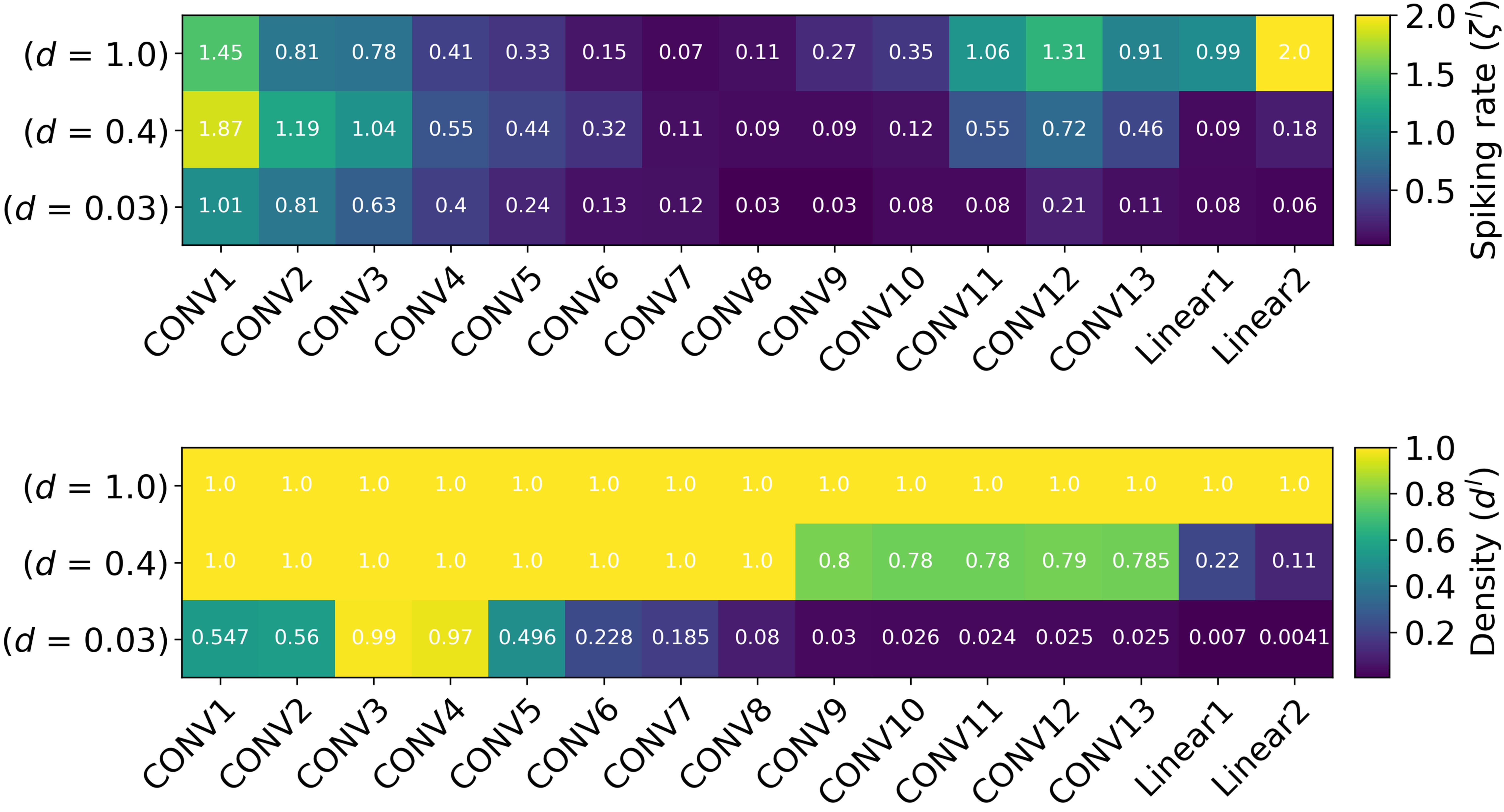}
\centering
   \caption{Heatmap visualization of spiking rate (activity) and model parameter density per layer for three different values of $d$ with VGG16 on CIFAR-10.}
\label{fig:density_vs_spike_rate_heatmap}
\vspace{-5mm}
\end{figure}

\subsection{Analysis of Energy Consumption}
\label{subsec:energy_analysis}

\subsubsection{Spiking Activity}
\label{subsub:spiking_act} 
Based on the assumption that a generated SNN spike consumes a fixed amount of energy \cite{dsnn_conversion1}, earlier works \cite{rathi2020iclr, dsnn_conversion_abhronilfin} have adopted the average spiking activity (also known as average spike count) of an SNN layer $l$, denoted ${\zeta}^l$, as a measure of layer computation energy of the model. More specifically, ${\zeta}^l$ is computed as the ratio of the total spike count over $T$ steps over all the neurons of the layer $l$ to the total number of neurons in that layer. Thus, lower spiking activity reflects higher compute efficiency.   

\textbf{Observation 4:} \textit{Model compression generally reduces spiking activity of layers which have lower pruning sensitivity.\footnote{We define a layer's pruning sensitivity via its non-zero weight parameter density. For a given target global density $d$, we
identify a layer $l$ that has higher layer density $d^l$ to have higher sensitivity.}}

As exemplified for VGG16 on direct input CIFAR-10 illustrated in Fig. \ref{fig:density_vs_spike_rate_heatmap}, we obtained reduced spiking activity for models with aggressive irregular pruning, especially at later layers. This may be related to the density of those layers. Thus we conjecture that layers that have lower pruning sensitivity have reduced spiking activity. In contrast, initial layers have comparatively higher parameter density and less differences
in spiking activity between compressed and uncompressed models.     

\begin{figure*}[!t]
\includegraphics[width=0.90\textwidth]{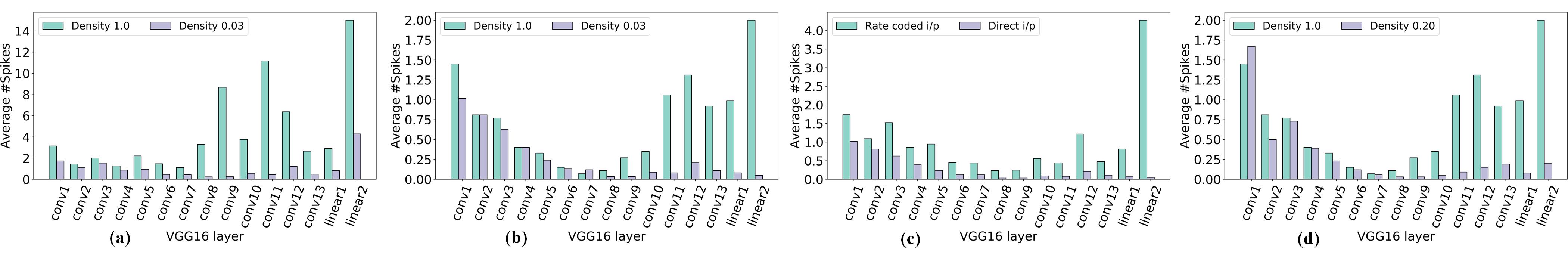}
\centering
   \caption{(a), (b) show average spiking activity plots for different layers of VGG16 for different densities with rate coded and direct inputs, respectively. (c) Comparison of spiking activity for rate coded and direct input for model of fixed target density $d = 0.03$. (d) Average spiking activity for different densities with channel pruning. Note that for both rate coded and direct input models to have comparable accuracy on CIFAR-10 dataset, we 
   compute the per layer spiking activity averaged over one mini-batch of 32 images.}
\label{fig:spike_vs_model_comparison}
\vspace{-2mm}
\end{figure*}

As shown in Fig. \ref{fig:spike_vs_model_comparison}(a)-(b), the per-image average number of spikes at each layer of a compressed ($d$ = $0.03$) VGG16 over the reduced time steps is smaller than an uncompressed variant while classifying both rate coded and direct input variants of CIFAR-10. Fig. \ref{fig:spike_vs_model_comparison}(c) depicts the benefits of low latency direct input coding over comparatively higher latency rate coded inputs, achieving significantly lower spike activity for almost all layers. 
Note that, in Fig. \ref{fig:spike_vs_model_comparison}(c) we observe reduced spike activity of up to $85.4\times$ (linear layer 2). On the contrary, reduction at the initial layers can be as little as $\mathord{\sim}1.36\times$ (CONV 3), which we conjecture is due to higher spiking activity at the initial layers as a result of the direct coded inputs. We observe a similar reduction in average spike count for channel pruned SNN models (see Fig. \ref{fig:spike_vs_model_comparison}(d)).

\subsubsection{FLOPs and Computation Energy}
\label{subsub:flops_compute_energy}

\begin{figure}[!t]
\includegraphics[width=0.36\textwidth]{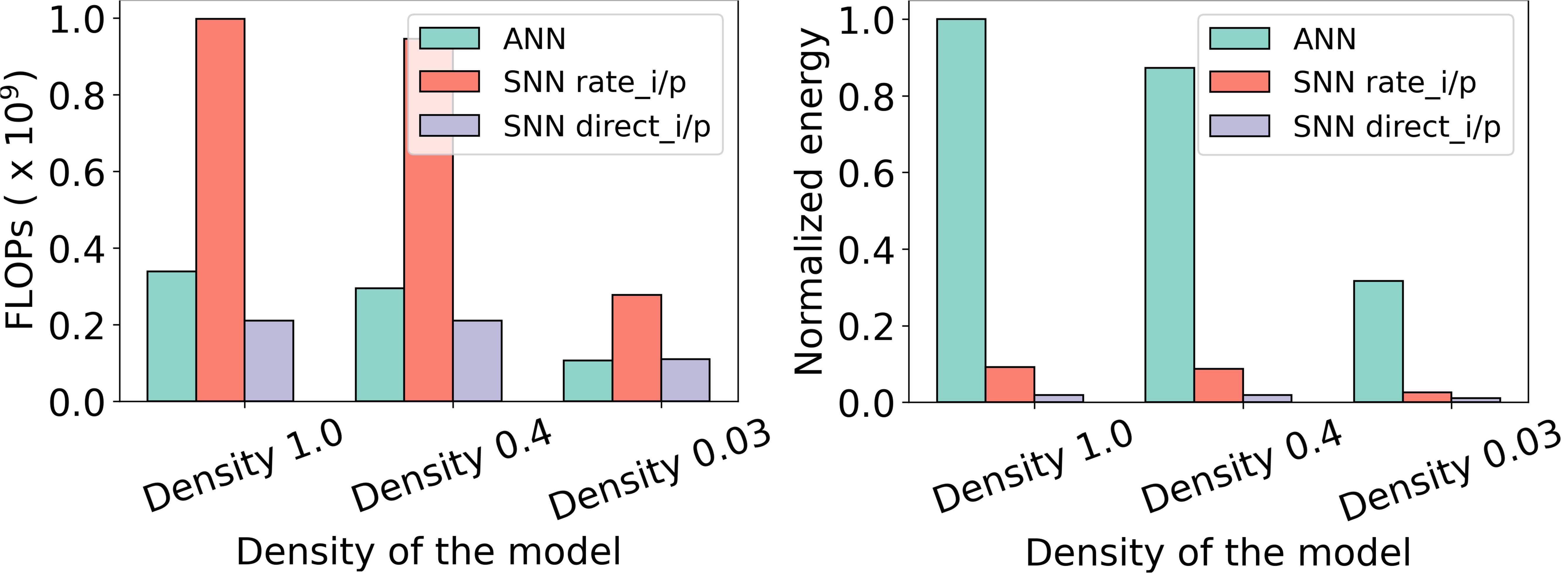}
\centering
\caption{Comparison of FLOPs and computation energy of VGG16 between ANN and SNN variants having both direct and rate coded inputs while classifying CIFAR-10.}
\label{fig:FLOPs_and_energy_plot_VGG16_c10}
\vspace{-2mm}
\end{figure}

Table \ref{tab:flops} shows the FLOPs computation equations for a convolutional layer $l$ with weight tensor $\textbf{W}^l$ and input activation tensor $\textbf{A}^l$, producing an output activation map $\textbf{O}^l \in \mathbb{R}^{H_{o}^l \times W_{o}^l \times C_{o}^l}$ for uncompressed and compressed variants of ANNs, denoted as $FL^l_{ANN}$ and $FL^l_{ANN_c}$, respectively. Here $d^l$ is the parameter density of CONV layer $l$. For SNNs having significant activation sparsity, we compute the $FL^l_{SNN}$ and $FL^l_{SNN_c}$ corresponding to uncompressed and compressed variants, respectively.\footnote{For details of the computation refer to \cite{Kundu2021WACV}.} Based on this formulation we now write the convolutional computation energy of SNNs with direct input, $E_{SNN^d_C}$ as,

{\small
\vspace{-4.1mm}
\begin{align}
    E_{SNN^d_C} &=(FL^1_{ANN_C})\cdot{E_{MAC}}+(\sum_{l=2}^{L}FL^l_{SNN_C})\cdot{E_{AC}} 
\end{align}
}

\noindent
where $E_{MAC}$ and $E_{AC}$ are the energy consumption of a MAC and AC operation, respectively, and for a $45 nm$ CMOS technology $E_{MAC}$ consumes $\mathord{\sim}32 \times$ more energy than $E_{AC}$ \cite{horowitz20141}. Although the first layer requires MAC operations to support direct inputs, its contribution to the total energy consumption remains typically negligible. Moreover, it can be further reduced by adjusting the baseline model to have lower $C^1_o$. Also due to  reductions in inference latency, the SNNs with direct input coding require lower FLOPs and achieve higher compute efficiency than their rate coded counterparts.
As exemplified in Fig. \ref{fig:FLOPs_and_energy_plot_VGG16_c10}, VGG16 with direct input and $d=0.4$ can have an improved FLOPs and compute efficiency of up to $\mathord{\sim}4.5 \times$ compared rate coded variants. Also, compared to an uncompressed and iso parameter ANNs, direct input coded VGG16 have $\mathord{\sim}98 \times$ and $\mathord{\sim}47 \times$  higher compute efficiency, respectively.

\begin{table}[!ht]
\scriptsize\addtolength{\tabcolsep}{-6pt}
\begin{center}
{\makegapedcells
\begin{tabular}{|c|c|c|}
\hline
{Model} & \multicolumn{2}{|c|}{FLOPs of a CONV layer $l$}\\
\cline{2-3}
        & Variable & Value \\
\hline
{$ANN$} &  $FL_{ANN}^l$ & $(k^l)^2\times H_o^l\times W_o^l\times C_o^l\times C_i^l$\\
\hline
{$ANN_c$} \cite{kundu2020pre} &  $FL_{ANN_C}^l$ & $(k^l)^2\times H_o^l\times W_o^l\times C_o^l\times C_i^l\times d^l$\\
\hline
{$SNN$} & $FL_{SNN}^l$ & $(k^l)^2\times H_o^l\times W_o^l\times C_o^l\times C_i^l \times {\zeta}^l$ \\
\hline
$SNN_C$ & $FL_{SNN_C}^l$  & $(\sum_{x=0}^{H_o^l -1}\sum_{y=0}^{W_o^l -1} \sum_{p=0}^{C_o^l -1}\sum_{n=0}^{C_i^l -1}\sum_{i=0}^{k^l-1}\sum_{j=0}^{k^l-1}$ \\
  &  &    $\zeta_{n,{x+i},{y+j}}^l \times  
        m_{p,n,i,j}^l)$ \\
\hline
\end{tabular}}
\end{center}
\caption{Convolutional layer FLOPs for ANN and SNN models}
\label{tab:flops}
\end{table}
\vspace{-0.4cm}
\section{Conclusions and Broader Impact}
\label{sec:concl}
This paper proposes a direct input driven hybrid sparse-learning approach to yield compressed deep SNN models that have high energy efficiency and reduced inference latency. In particular, we first use a novel attention-guided ANN compression, then convert the ANN to an SNN by sequentially fixing the firing threshold of each layer, and finally training the SNN using a sparse-learning based approach that starts with the compressed ANN weights. Our framework supports both irregular and channel pruning, allowing the generated models to speed up inference on a broad range of computing platforms and edge devices. Our experimental evaluations showed that the generated SNNs can have compression ratios of up to $33.4\times$ with negligible drop in accuracy. Moreover, compared to the rate coded input driven models we can achieve up to an order of improved inference latency. We believe this along with the potential speed up in per layer compute time for channel pruned models will help us take one step forward in making SNNs a promising low-power application where ultra fast inference speed is a necessity.


%

\ifCLASSOPTIONcaptionsoff
  \newpage
\fi




{\small
\bibliographystyle{IEEEtran}
\bibliography{biblio}
}

%



%
\vspace{-11mm}
\end{document}